\documentclass[10pt,twocolumn,letterpaper]{article}

\usepackage{cvpr}              

\usepackage{graphicx}
\usepackage{amsmath}
\usepackage{amssymb}
\usepackage{booktabs}
\usepackage{multirow}
\usepackage{multicol}
\usepackage{makecell}
\usepackage{color,soul}

\usepackage[pagebackref,breaklinks,colorlinks]{hyperref}

\usepackage[capitalize]{cleveref}
\crefname{section}{Sec.}{Secs.}
\Crefname{section}{Section}{Sections}
\Crefname{table}{Table}{Tables}
\crefname{table}{Tab.}{Tabs.}

\def\cvprPaperID{5514} 
\def\confName{CVPR}
\def\confYear{2022}

\begin{document}

\title{Towards Benchmarking and Evaluating Deepfake Detection}

\author{Chenhao Lin\\
Xi'an Jiaotong University\\
{\tt\small linchenhao@xjtu.edu.cn}
\and
Jingyi Deng\\
Xi'an Jiaotong University\\
{\tt\small misscc320@stu.xjtu.edu.cn}
\and
Pengbin Hu\\
Xi'an Jiaotong University\\
{\tt\small hupb666@stu.xjtu.edu.cn}
\and
Chao Shen\\
Xi'an Jiaotong University\\
{\tt\small chaoshen@mail.xjtu.edu.cn}
\and
Qian Wang\\
Wuhan University\\
{\tt\small qianwang@whu.edu.cn}
\and
Qi Li\\
Tsinghua University\\
{\tt\small qli01@tsinghua.edu.cn}
}
\maketitle

\begin{abstract}
   Deepfake detection automatically recognizes the manipulated medias through the analysis of the difference between manipulated and non-altered videos. 
It is natural to ask which are the top performers among the existing deepfake detection approaches to identify promising research directions and provide practical guidance.
Unfortunately, it's difficult to conduct a sound benchmarking comparison of existing detection approaches using the results in the literature because evaluation conditions are inconsistent across studies.
Our objective is to establish a comprehensive and consistent benchmark, to develop a repeatable evaluation procedure, and to measure the performance of a range of detection approaches so that the results can be compared soundly. 
A challenging dataset consisting of the manipulated samples generated by more than 12 different methods has been collected, and 11 popular detection approaches (9 algorithms) from the existing literature have been implemented and evaluated with 6 fair-minded and practical evaluation metrics. Finally, 92 models have been trained and 644 experiments have been performed for the evaluation.
The results along with the shared data and evaluation methodology constitute a benchmark for comparing deepfake detection approaches and measuring progress. 
\end{abstract}

\section{Introduction}
\label{sec:intro}

The recent emergence of face manipulation technology based on deep learning, also known as Deepfake, misleads people into believing the fake words and deeds, posing a new threat to violation of privacy, identity, financial, legal, and even national security~\cite{chesney2019deep}. As the deepfake videos and pictures spread, deepfake detection techniques have been increasingly focused on to prevent emerging malicious face manipulation threats. 

Along with the emergence of large-scale deepfake forensic datasets, some forgery detection benchmarks have been established recently. Unfortunately, they seldom focus on fair and comprehensive evaluation of existing state-of-the-art deepfake detection approaches.
The most obvious problem of existing benchmarks is that although they evaluate detection methods on consistent evaluation datasets, the benchmarking models are usually trained on different data, which leads to an unfair comparison. In other words, the evaluation performances of detection models are highly related to their training data and models trained on data with larger scale, higher quality and more diversity of manipulation approaches are able to attain better evaluation performance.

In addition, as for benchmark datasets, they are usually compared in terms of data scale and diversity of manipulation and perturbation. However, there is no quantitative comparison indicating which dataset is more challenging and appropriate to train a robust deepfake detection model. 
As a result, there is an urgent need to construct a systematic, fair and consistent benchmark with practical evaluation metrics to identify current achievements and perceive the future requirements in deepfake detection.

Although several advanced deepfake detection methods have been proposed and proved to be effective, it is difficult to soundly quantify the contribution of existing works due to the following reasons.
Firstly, unlike popular image classification and object detection tasks usually using same datasets for training to ensure a fair comparison, many deepfake detection methods are trained on different datasets but evaluated on the same test data.
For example, many existing works~\cite{zhao2021multi,dang2020detection} directly apply publicly available trained models instead of re-implementing these methods using the same training data for the evaluation. Such inconsistent training data leads to unfair evaluations and comparisons of the existing detection methods and therefore it is difficult to measure whether the performance contributions are brought by the method itself or its adopted training data\cite{baek2019wrong, brigato2021tune}. 

Secondly, most deepfake detection methods with outstanding performance can be impractical due to the overfitting problem and poor transferability, since many of them are trained and evaluated on the same domain distributed datasets containing limited manipulation methods.
Predictably, the detection performance of these methods may experience a significant decrease when testing on the datasets with different distributions or deploying in realistic scenarios dealing with fake data generated by different manipulation approaches ~\cite{li2020face,chai2020makes}.

Thirdly, the widely used evaluation metrics, including AUC (area under the ROC curve) and accuracy, are not sufficient to reflect a comprehensive performance of detection methods. The essential and practical evaluation metrics including time and space complexity
have drawn nil attention in previous research, which can result in the low efficiency of top detection methods for large-scale forged videos or images in realistic scenarios.


To address these limitations, this paper proposes a fair, comprehensive and strict benchmark on our collected standard datasets and our proposed Imperceptible and Diverse test (ID test) set. The standard datasets is constructed by incorporating several existing representative deepfake forensic datasets and used to train and evaluate popular deepfake detection methods re-implemented during the experiment. To better simulate a realistic media environment, an Imperceptible and Diverse test (ID test) set has been proposed, containing hard and diverse samples selected from public datasets and our hosted private dataset, and merely used for the evaluation. The forged videos in ID test set are highly indistinguishable to both human eyes and detection algorithms, synthesised by various classic manipulation approaches and distorted by commonly encountered perturbations.
In addition to using AUC and accuracy metrics for evaluation, four complementary evaluation metrics have been applied to measure the benchmarking methods from different aspects, including forgery detection ability, robustness, efficiency, and practicability. 
To guarantee the completeness of the experiment, the deepfake datasets were initially categorized according to their adopted forgery methods and then experiments were conducted on intra- and inter-class evaluations.

Through a comprehensive and quantitative analysis of the results from 644 evaluation experiments, this work presents several important findings.
First, 
the forgery detection ability of all 11 popular deepfake detection approaches drops significantly on a realistic and challenging dataset, indicating the performance fails to satisfy the requirement for real-world applications.
Second, we find the overall performance of popular detection methods shows no significant difference under strictly uniform evaluation conditions, unlike the claim in previous studies that a detection method is significant better than another by using a specific evaluation configuration.
Third, taking into account detection ability, generalization, robustness and practicability simultaneously, no one method shows comprehensive superiority over others. 
Considering from different perspectives, Multiple-attention achieves the best AUC in terms of detection performance with a high time complexity, Patch-Xception-Block2 and Patch-Resnet-Layer1 has a relatively high detection ability with very low inference time and memory consumption. Conv LSTM owns the top-performing generalization ability with the highest time complexity.

\begin{table*}[h!]
\centering
\begin{tabular}{l|c|c|c|c|c|c|c|c|c}
\hline
\multirow{2}{*}{Dataset}  & 
\multicolumn{2}{c|}{Total Frames/Images} & 
\multirow{2}{*}{\makecell{Video-level \\ Split}} & 
\multicolumn{3}{c|}{Manipulation Method} & 
\multirow{2}{*}{\makecell{Extra \\ Test Set}}& \multirow{2}{*}{\makecell{Perturb}} &
\multirow{2}{*}{\makecell{Bench-\\mark}}\\
&Real&Fake&& AE & GAN & Graphic & & & \\
\hline\hline
UADFV & 17,329 & 16,991 & 3:1:1 & 1 & --  & -- & -- & -- &  13\\
DF-TIMIT & 34,003 & 34,023 & 3:1:1 & -- &  1 & -- & -- & -- & 11\\
Celeb-DF-v2 & 358,790 & 2,116,768 & 13:1:1 & 1 & -- & -- & -- & -- & 7\\
DeeperForensics-1.0 & 509,128 & 508,944 & 7:1:2 & 1 & -- & -- & \checkmark & 7 & 7\\
FaceForensics++ & 509,914 & 1,321,408 & 5:1:1 & 1 & 1 & 1 & -- & -- & 23 \\
DFDC & 5,635,501 & 29,075,744 & 5:1:1 & 2 & 3 & 1 & \checkmark & 19 & 2119\\
ForgeryNet & 2,848,548 & 1,054,671 & 48:3:7 & 1 & 2 & - & \checkmark & 36 & 11 \\
\hline
\end{tabular}
\caption{Overview of popular forensic datasets. We list the most crucial information reflecting the pros and cons of datasets to train and evaluate deepfake detection methods. Dataset scale and data distribution are reflected by the measurement of total frames. 
Dataset quality is measured by the diversity of manipulation methods and perturbation types, as well as the inclusion of hard examples. Dataset popularity is quantified by the number of benchmarks it has been established on.}
\label{tab:datasets}
\end{table*}

\section{Deepfake Creation and Detection}
This paper concentrates on integrating the current popular forensic datasets and detection methods to release a comprehensive benchmark.
We firstly briefly introduce the popular deepfake creation methods and corresponding datasets. Then we summarize and roughly classify the existing deepfake detection approaches into four categories based on their strategies. Last subsection describes the existing evaluation metrics and forensic benchmarks. 

\subsection{Deepfake Creation and Forensic Datasets}

\subsubsection{Deepfake Creation}
Existing popular face manipulation approaches can be roughly classified into three types.

\textbf{\emph{Autoencoder-based Manipulation}} maintains an autoencoder for each pair of face swapped identities. The autoencoder consists of a shared encoder that enables encoding common features of both identities and a specific decoder to reconstruct the source face in the target image.
As one of the most widely used manipulation types, majorities of forensic datasets contain this type of manipulated data. For example, UADFV~\cite{li2018ictu} uses an autoencoder-based software, FakeApp~\cite{FakeApp} 
to generate 49 fake videos. FaceForensics++~\cite{rossler2019faceforensics++} applies an open-source tool called FaceSwap~\cite{FaceSwap} 
to manipulate 1,000 videos. DeeperForensics-1.0~\cite{jiang2020deeperforensics} designs a novel autoencoder-based framework, DF-VAE, to manipulate 10,000 videos. Recently proposed  ForgeryNet leverages DeepFakes~\cite{perov2020deepfacelab} to generate thousands of videos and images. 

\textbf{\emph{GAN-based Manipulation}} leverages generators, which trained by the contest with discriminators, to generate whole source face or some source face attributes in manipulated face images. Benefiting from its powerful learning ability and manipulation performance, DeepFake-TIMIT~\cite{korshunov2018deepfakes} leverages a GAN-based face-swapping algorithm~\cite{faceswap-GAN} to synthesize 620 fake videos. FaceForensics++ tampers 1,000 videos by using a two-stage face swapping method named FaceShifter~\cite{li2019faceshifter} and ForgeryNet selects GAN-based FSGAN~\cite{nirkin2019fsgan} and FaceShifter~\cite{li2019faceshifter} to generate tens of thousands of videos and images.

\textbf{\emph{Graphic-based Manipulation}} forges face images by modeling source face landmarks and deforming it to match the landmarks of target images, which often followed by a blending operation~\cite{huang2012facial}. Within existing forensic datasets, FaceForensics++ adopts 3D-Faceswap~\cite{3d_FaceSwap} to manipulated 1,000 videos and DFDC uses MM/NN face swap method proposed in~\cite{huang2012facial} to forge parts of videos.

\subsubsection{Forensic Datasets}
According to our analysis and classification of deepfake creation approaches, existing forensic datasets commonly comprise of deepfake videos and images generated by limited types of manipulated approaches due to the high consumption of resources and time. For example, UADFV, DeepFake-TIMIT, Celeb-DF~\cite{li2020celeb}, DeeperForensics-1.0 and ForgeryNet are constructed by one or two types of forgery approaches. 
Only two well considered datasets, FaceForensics++ and DFDC, utilize complete forgery types to offer a more complex and practical dataset. The existing popular forensic datasets have been summarized and listed in Table \ref{tab:datasets}. For the datasets with multiple facial forgery types like deepfake, expression editing, and attribute manipulation, we list the information about fake data manipulated by deepfake and their related real data.

Additionally, to better simulate real-world scenarios and prevent detection bias, the newer proposed DeeperForensics-1.0, DFDC and ForgeryNet datasets introduce hidden/private test sets respectively, guaranteeing manipulation diversity and visual quality of fake videos. However, some shortages of these three test sets are unrevealed. 
DeeperForensics-1.0 carries out a user study to ensure the visual reality of the fake videos, while the data distribution and data diversity are unknown.
On the contrary, DFDC and ForgeryNet clearly describe the details of their hidden test sets, while the manipulated data are not manually checked to ensure their imperceptibility to human eyes.

\subsection{Forgery Detection}
\label{forgery_detection}

To defend the abused synthetic media, considerable efforts have been undertaken by researchers and communities these years, promoting the raise of multiple detection methods. We roughly categorize the existing forgery detection methods into  {\it intra-frame} detection and {\it inter-frame} detection with several subclasses.

\subsubsection{Intra-Frame Detection}

Intra-frame detection refers to the image-level detection, performing an image-level supervised binary classification on frame images. Its inference predictions contain image-level prediction and video-level prediction. Image-level prediction assigns a prediction score to each image, while video-level prediction aggregates the image-level scores by voting or averaging and assigns the aggregating score to each video.
According to the principle difference of detection methods, intra-frame detection can be further subdivided into {\it knowledge-driven detection}, {\it data-driven detection} and {\it multi-stream-driven detection}. 

\textit{\textbf{Knowledge-Driven detection}} incorporates the domain knowledge to pre-define an interpretable artifact generated by manipulation, as distinctive embeddings to train a model for classification. To ensure learning interpretability, such kind of detection methods often need to extract artifacts related features or generate artifacts-emphasized supervised information in the data pre-processing phase. For example, Headpose~\cite{yang2019exposing} extracts features of landmark difference, F3Net~\cite{qian2020thinking} extracts features of frequency difference, FWA~\cite{li2018exposing} generates training data with emphasized resolution inconsistency and Face X-ray~\cite{li2020face} generates masks representing blending boundary.

\textit{\textbf{Data-Driven detection}} relies on the large-scale dataset and the robust learning capability of DNN, aiming to adopt or design an efficient deep neural network to automatically learn the multi-scale difference features between real and fake images for classification~\cite{afchar2018mesonet,chai2020makes,cozzolino2018forensictransfer}. Typically, through such detection, the intuition behind the learning process and the distinguishable activated features are commonly analyzed.

\textit{\textbf{Multi-Stream-Driven detection}} is a fusion of the above two types of strategies. Such kind of methods usually develops multiple streams to concurrently learn interpretable features and latent features, and finally fuses multi-stream features to achieve forgery detection~\cite{zhou2017two,chen2019attention}.

\subsubsection{Inter-Frame Detection}

Inter-frame detection, also known as video-level detection, attempts to learn the temporal artifacts from sequential manipulated video frames. Most existing works adopt a CNN to extract per-frame features and then use recurrent neural networks to explore inter-frame inconsistencies~\cite{guera2018deepfake,sabir2019recurrent,montserrat2020deepfakes}. Moreover, some popular techniques for video classification, such as Long Short Term Memory (LSTM) and I3D~\cite{jiang2020deeperforensics}, are also adopted for deepfake detection.

\subsection{Evaluation Metrics and Forensic Benchmarks}

The most widely used evaluation metrics for deepfake detection are AUC~\cite{li2020face,yang2019exposing,li2018exposing}, precision~\cite{chai2020makes}, and accuracy~\cite{rossler2019faceforensics++}. Besides these classic evaluation metrics, DFDC puts forward Weighted PR, assigning a weight to false positives of precision due to the large class imbalance of fake and real videos in organic traffic. Although the above evaluation metrics are commonly used for evaluation, it is not enough to explore the comprehensive utility of detection methods in practical scenes, as the metrics only consider the prediction correctness but ignore other important practical evaluation criteria.


As for existing benchmarks, in addition to the aforementioned FaceForensics Benchmark and DFDC, FaceForensics++ evaluates six intra-frame forgery detection methods on its dataset. Celeb-DF evaluates ten methods using the publicly released pre-trained models. DeeperForensics-1.0 integrates five baselines including both intra-frame and inter-frame methods. However, most of these benchmarks fail to fairly evaluate existing deepfake detection methods by re-implementing these methods with same training data, which introduces an evaluation bias.

\section{Evaluation Methodology}

As face manipulation is being increasingly easy to access by mature algorithms and even off-the-shelf software, practical forensic approaches are in desperate need. 
However, among existing popular deepfake detection methods, it is difficult to measure which one is more applicable when facing real threats due to the lack of fair and proper evaluated benchmark. To address this issue and further promote the research in this field, we propose a fair, comprehensive, and strict benchmark by integrating 7 popular forensic datasets and 11 representative forgery detection methods. At the same time, we additionally apply 4 complementary curves to thoroughly evaluate their robustness and practicability. According to our categorization, this benchmark investigates the detection capabilities of each type of state-of-the-art forensic methods against different types of manipulations. What's more, we 
perform comprehensive experiments on our ID test set to explore their generalization capabilities in realistic scenarios.

\subsection{Evaluation Datasets and Algorithms}

\subsubsection{Standard Datasets}

The standard datasets consist of real and manipulation data from autoencoder-based UADFV, Celeb-DF, DF-1.0, GAN-based DF-TIMIT(higher quality), and mixed-manipulation-based FF++(Raw), DFDC and ForgeryNet. The data scales extracted from these 7 datasets are proportional to reflect their original scale difference.
Specifically, taking the frame number of 2,527,384 frames in FaceForensics++ as a baseline, the extracted frame number from other datasets doubles or decreases.
Each dataset in the standard datasets was split to training, validation and test to perform method re-implementing and evaluation. Specifically, the video-level split of each dataset complies with the default dataset setting if it is released, instead, we carry out a reasonable split for the datasets as illustrated in Table \ref{tab:datasets}. The frame-level data adopted in experiments were randomly extracted from split videos and keep a frame-level split ratio of 14:1:1, following the split strategy of FaceForensics++. Moreover, we maintain the distribution of real and fake data of each dataset while balance them in experiments.

\subsubsection{Imperceptible and Diverse Test (ID Test) Set}

To explore the robustness of forensic approaches when confronting the threats of fake videos with high visual authenticity and rich content diversity, we construct a high-quality Imperceptible and Diverse test (ID test) set by integrating the hard (high imperceptible) examples from our benchmarking 7 public datasets and our hosted private dataset. 
The hard examples from public datasets undergo a two-phase selection pipeline, namely detection model selection and user perception selection. 
The detection model selection retains falsely accepted fake examples with high confidence. Then the user perception selection carries out a blind experiment and preserves the high-quality fake videos considered real by 15 out of 30 participants, which means that these examples are indistinct to both detection models and human visions.
The hard examples from our private dataset are our self-generated fake examples manipulated by recent introduced GAN-based FSGAN~\cite{nirkin2019fsgan} and autoencoder-based MegaFS~\cite{zhu2021one} approaches, of which the original data are images of CelebA~\cite{liu2015deep} and raw videos of FaceForensics++. 
These manipulated hard examples also go through the selection pipeline to guarantee their visual reality and finally 40 videos manipulated by FSGAN and 2937 images manipulated by MegaFS are preserved. Overall, ID test set comprises 976 fake videos and 2348 real videos, from which 25,697 fake images and 25,697 real images are extracted, respectively.

\begin{table}[h!]
\begin{center}
\begin{tabular}{l|c|c|c|c}
\hline
& 
AE & 
GAN & 
Graphic&
Unknown\\
\hline\hline
Video & 522 & 202 & 22 & 230\\
Image & 10,514  & 10,778 & 2,171 & 2,234\\
\hline
\end{tabular}
\caption{Overview of manipulation type distribution of ID test set. The unknown manipulation type indicates that the related information of these data were unavailable from its source dataset.}
\label{tab:idtestset_distribution}
\end{center}
\end{table}

\begin{figure}[t]
\vspace*{-5mm}
\begin{center}
   \includegraphics[width=\linewidth]{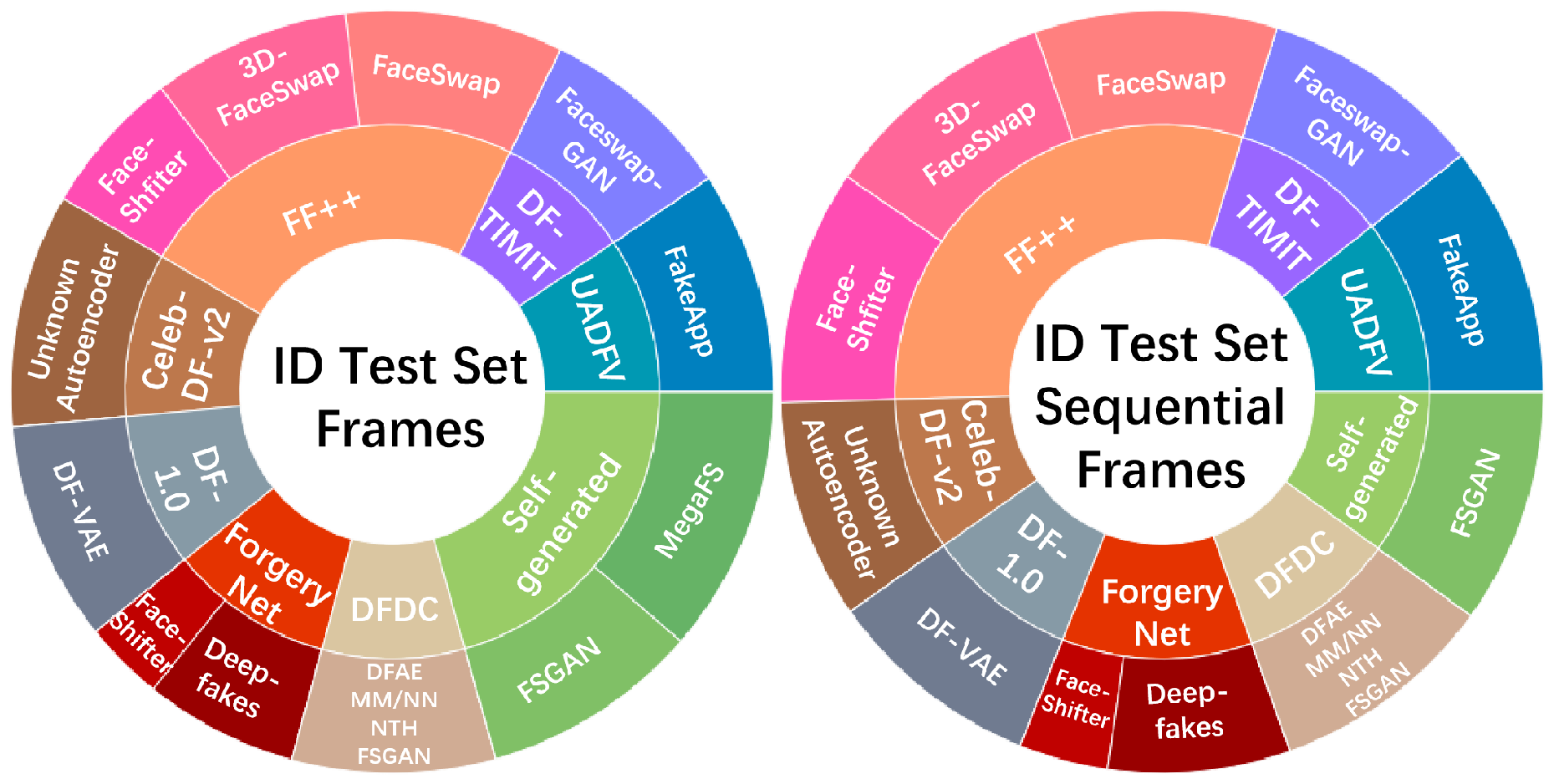}
\end{center}
   \caption{Illustration of frame data and sequential frame data distribution of each manipulation methods in ID test set.}
   \vspace*{-5mm}
\label{fig:id_test_set_dist}
\label{fig:onecol}
\end{figure}

\begin{figure}[t]
\vspace*{-1mm}
\begin{center}
   \includegraphics[width=0.9\linewidth]{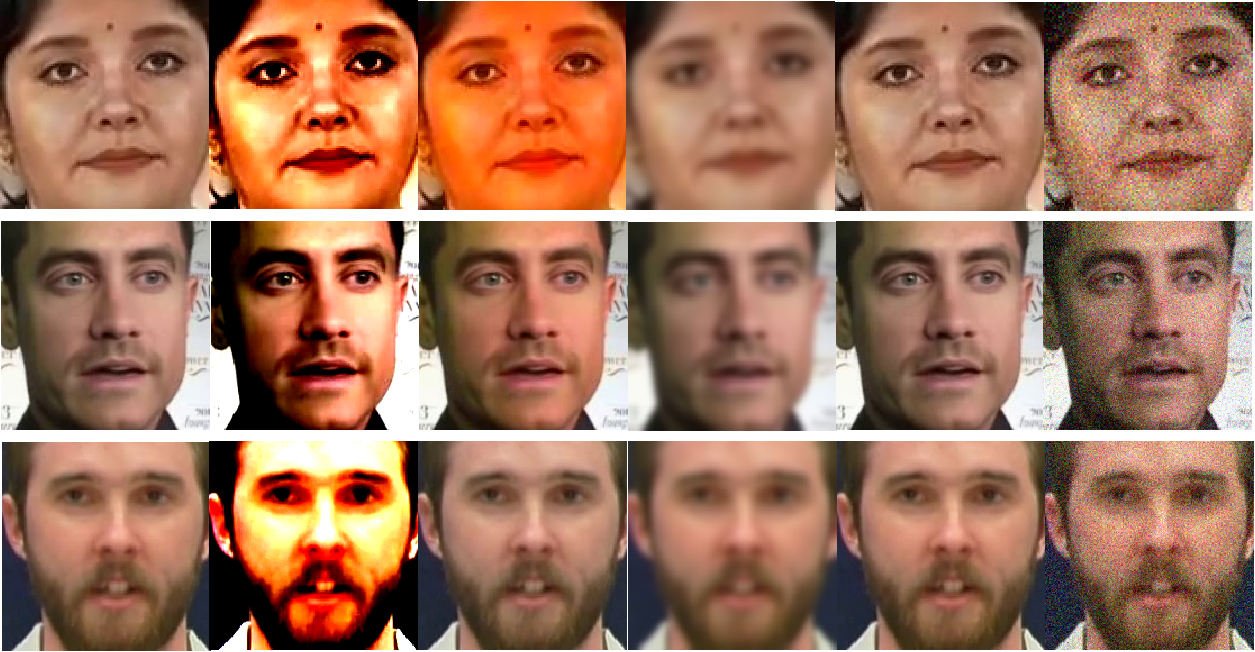}
\end{center}
   \caption{Illustration of face images in ID test set with different perturbations. From left to right are raw face images and face images with color contrast change perturbation, color saturation change perturbation, Gaussian blur perturbation, JPEG compression perturbation and white Gaussian noise perturbation.}
\label{fig:perturbation}
\label{fig:onecol}
\end{figure}

To guarantee the diversity, fake data in ID test set achieves full coverage of manipulation types and at least 13 manipulation approaches. The specific manipulation type distribution and forgery approach distribution of ID test set are shown in Table \ref{tab:idtestset_distribution} and Figure~\ref{fig:id_test_set_dist}. We extract almost equal quantity of frames and sequential frames of each manipulation method, which enables fair evaluation of image-level and video-level detection methods trained by different datasets.
Moreover, to simulate the restricted video quality caused by the video pre-processing pipeline, 5 types of common perturbations, as shown in Figure~\ref{fig:perturbation}, are added to videos and images for extra evaluation.

\begin{table*}[h!]
\begin{center}
\begin{tabular}{l|c|c|c|c|c|c|c}
\hline
Methods  & 
Category &
OS & 
Data Pre &
Backbone/Method & 
Param(M)& 
GFLOPs&
Infer T(ms)\\
\hline\hline
HeadPose & Frame-Know & yes & no & SVM & - & - & 159.70\\
FWA-Resnet50 & Frame-Know & part & D & Resnet50 & 25.56 & 8.24 & 101.21\\
Face X-ray& Frame-Know  & no & D & HRNet-W48-C & 77.47 & 42.58 & 35.62\\
Xception & Frame-Data & part & D & XceptionNet & 20.81 & 16.84 & 5.25\\
Meosonet-4 & Frame-Data & part & D + A & 4-layer Conv & 0.28 & 0.12 & 5.11\\
MeosoInception-4 & Frame-Data & part & D + A & 2-Inception+2-Conv & 0.28 & 0.11 & 7.31\\
Patch Resnet Layer1 & Frame-Data & yes & D + A & Resnet18 &  0.15 & 2.10 & 0.73 \\
Patch Xception Block2 & Frame-Data & yes & D + A & XceptionNet & 0.19 & 3.34 & 1.17\\
FFD & Frame-Know & yes & D & XceptionNet + Reg. Map & 20.82 & 16.84 & 6.04\\
Multiple-attention & Frame-Know & part & D+A & EfficientNet-b4& 18.83 & 6.80 & 25.48\\
Conv LSTM & Video & no & D & InceptionV3 \& LSTM & 30.36 & 229.48 & 221.64 \\
\hline
\end{tabular}
\caption{Overview of our evaluated forensic detection algorithms. OS represents their open source situation. Data Pre is data pre-processing procedure, in which D stands for face detection and A stands for face alignment. Infer T represents inference time.
}
\label{tab:detection_methods}
\end{center}
\end{table*}

\subsubsection{Forensic Detection Algorithms}

We evaluate 11 declared state-of-the-art forensic detection approaches (9 algorithms) covering the categories described in \textit{Forgery Detection}, Table \ref{tab:detection_methods}. Within the intra-frame detection category, we select knowledge-driven Headpose~\cite{yang2019exposing}, Face X-ray~\cite{li2020face}, FWA~\cite{li2018exposing}, and data-driven Xception~\cite{rossler2019faceforensics++}, Mesonet~\cite{afchar2018mesonet}, MesoInception4~\cite{afchar2018mesonet}, Patch-forensics~\cite{chai2020makes}, FFD~\cite{dang2020detection} and Multiple-attention~\cite{zhao2021multi}. 
Among inter-frame detection, we evaluate Convolutional LSTM~\cite{guera2018deepfake}. For the methods with multiple backbones, we attach the backbone gained best reported performance.

\textit{\textbf{Pre-processing.}}
Common data pre-processing pipeline includes frame extraction, face cropping and face alignment. The image processing toolkit Dlib~\cite{king2009dlib} is applied to do the face detection and alignment. For the detection algorithms with additional data pre-processing, including HeadPose, FWA, Face X-ray and FFD, we follow their implementations and attach the specific implementation details in the Appendix A.


\textit{\textbf{Implementation Details.}}
We re-implement most algorithms without released training codes and directly adopt the source code for the four open-sourced algorithms. 
Considering fair comparison, for each experiment, we re-train all these algorithms on the same training set and optimize the parameters according to the dataset scale. The re-training process completely refer to the original proposed papers.  
To ensure the implementation correctness, we also conduct verification experiments to show the consistency between our results and the reported results in the original papers. The comparative results are attached in the Appendix B.

\subsection{Proposed Evaluation Metrics}

In response to the inconsiderate limitation of commonly adopted evaluation metrics, we apply four complementary correlation plots to comprehensively analyze the robustness, practicability and efficiency of forensic classifiers.

\textit{\textbf{AUC versus Perturbation.}} The AUC/perturbation trade-off shows the robustness of forensic classifiers when facing the challenge of different types and extents of perturbations in realistic scenes. As shown in Figure~\ref{fig:perturbation}, we have applied five types of perturbation on our ID test set. 
The videos are distorted by each type of perturbation with random intensity, which is constrained within a rational range to guarantee the visual effect of pristine data.

\textit{\textbf{AUC versus FLOPs.}} Maintaining a high AUC score with low FLOPs indicates that the model owns an excellent forgery detection ability and consists of an optimized architecture with low computational complexity. Since efficient computing power is commonly required in the actual application environment, this AUC/FLOPs trade-off provides insights into the practicability of forensic classifiers.  

\textit{\textbf{AUC versus Number of Parameters.}} The correlation between AUC and number of parameters provides a deeper understanding of practicability. Outperformed classifier with less number of parameters requires less memory consumption, which is a crucial measurement in practical scenes.

\textit{\textbf{AUC versus Inference Time.}} The AUC/inference time trade-off reveals the efficiency of deploying forensic classifiers for testing a single image.

\subsection{Environment}

For fair evaluation, we conduct all the experiments and evaluate the benchmarking algorithms in a uniform environment, NVIDIA 2080Ti GPU and 128GB of RAM.

\begin{table*}
\begin{center}
\begin{tabular}{l@{\hspace{5pt}}c@{\hspace{5pt}}c@{\hspace{5pt}}c@{\hspace{5pt}}c@{\hspace{5pt}}c@{\hspace{5pt}}c@{\hspace{5pt}}c@{\hspace{5pt}}c@{\hspace{5pt}}c@{\hspace{5pt}}c}
\hline
Train  & UADFV & DF-TIMIT & Celeb-DF & DF-1.0 & FF++/DF & FF++/FS & FF++/FShifter& DFDC & ForgeryNet & \multirow{2}{*}{\makecell{Average \\ AUC\%}}\\
Test  & UADFV & DF-TIMIT & Celeb-DF & DF-1.0 & FF++/DF & FF++/FS & FF++/FShifter & DFDC & ForgeryNet \\
\hline\hline
Face X-ray& 97.1 & 98.5	& 97.8 & 84.7 & 99.8 & 99.8	& 99.7 & - & - & 96.7(96.7)\\
FWA-Resnet50& 57.3 & 99.1 & 60.3 & 60.5 & 80.6 & 61.2 & 50.0 & 47.6 & 50.3 & 63.6(67.0)\\
HeadPose & 88.3 & 62.3 & - & 57.2 & 57.0 & 52.3 & 61.2 & - & - & 63.0(63.0)\\
Meosonet-4 & 97.7 & 100.0 & 99.0 & 100.0 & 99.1 & 99.4 & 99.6 & 93.8 & 71.2 & 95.5(99.2) \\
MeosoIncept-4 & 97.8 & 100.0 & 98.7 & 100.0 & 98.9 & 96.8 & 99.5 & 94.8 & 67.0 & 94.8(98.8)\\
Patch-resnet & 98.4 & 100.0 & 89.8 & 99.9 & 99.5 & 99.7 & 99.5 & 90.1 & 60.5 & 92.4(98.1)\\
Patch-xception & 97.4 & 100.0 & 90.9 & 100.0 & 99.9 & 99.6 & 99.7 & 92.8 & 60.1 & 93.3(98.2)\\
Xception & 95.3 & 100.0 & 84.7 & 98.6 & 99.3 & 98.4 & 99.3 & 79.7 & 64.4 & 91.0(96.5)\\
FFD & 99.1 & 100.0 & 99.5 & 75.9 & 99.7 & 99.6 & 100.0 & - & - & 96.2(96.2)\\
Multi-attention & 97.6 & 100.0 & 99.9 & 99.9 & 99.2 & 99.7 & 99.6 & 99.1 & 80.0 & 97.2(99.4)\\
Conv LSTM & 99.8 & 100.0 & 91.9 & 99.9 & 99.9  & 98.7 & 100.0 & 81.4 & 64.5 & 92.9(98.6)\\
\hline
\end{tabular}
\caption{Results of forgery detection ability of different detection methods measured by frame-level AUC. We report the evaluation results of each method tested on the test data of each dataset. The results of Face X-ray and FFD on DFDC and ForgeryNet are unavailable due to the unavailability of additional supervised information without knowledge of mapping relation between source and target videos. The results of HeadPose on Celeb-DF-v2, DFDC and ForgeryNet are not reported because it is difficult to train SVM models on large-scale datasets. 
Patch-resnet and patch-xception refer to patch-resnet-layer1 and patch-xception-block2. 
FF++/FShifter refers to FF++/FaceShifter. 
The last column calculates the average AUC of methods, in which the results without parentheses are the average AUC across all datasets and the results within parentheses are that across commonly trained datasets excluding DFDC and ForgeryNet.}
\label{tab:forensic_ability}
\end{center}
\end{table*}

\section{Evaluation Results and Discussions}

We respectively present the evaluation results of the forgery detection ability, generalization ability, detection robustness, practicability, and efficiency/effectiveness trade-off of our benchmarking forgery detection methods, showing in the following subsections. In addition to the evaluation of forgery detection ability, all the results are evaluated on our ID test set. Due to the space restriction, for the evaluation of detection robustness, practicability and efficiency, we illustrate the representative experimental results in this section. 

\subsection{Evaluation Results of Forgery Detection Ability}
\label{Evaluation Results of Forgery Detection Ability}

We illustrate the forgery detection ability of our benchmarking methods in this section. For each experiment, all the detection methods are trained and evaluated on the same domain distributed data and measured by frame-level AUC. This evaluation demonstrates their detection ability for different kinds of manipulations. As shown in Table \ref{tab:forensic_ability}, most approaches can achieve superior performance when the distribution of training and test set are in the same domain. Multiple-attention gains the best average AUC score of 97.2\%(99.4\%), indicating its outstanding detection ability across different datasets. Models trained on DFDC and ForgeryNet are generally gain lower performance owning to the addition of new type of manipulation data or perturbation in their test dataset.

\begin{table*}
\begin{center}
\begin{tabular}{l@{\hspace{5pt}}c@{\hspace{5pt}}c@{\hspace{5pt}}c@{\hspace{5pt}}c@{\hspace{5pt}}c@{\hspace{5pt}}c@{\hspace{5pt}}c@{\hspace{5pt}}c@{\hspace{5pt}}c@{\hspace{5pt}}c}
\hline
Train  & UADFV & DF-TIMIT & Celeb-DF & DF-1.0 & FF++/DF & FF++/FS & FF++/FShifter & DFDC & ForgeryNet & \multirow{2}{*}{\makecell{Average \\ AUC\%}}\\
Test  & IDtest & IDtest & IDtest & IDtest & IDtest & IDtest & IDtest & IDtest & IDtest\\
\hline\hline
Face X-ray& 53.9 & 52.1 & 64.3 & 66.4 & 54.8 & 59.3 & 54.9 & - & - & 57.9(57.9)\\
FWA-Resnet50& 54.7 & 55.1 & 49.9 & 54.3 & 49.8 & 54.3 & 50.0 & 53.2 & 50.0 & 52.3(52.5)\\
HeadPose & 52.1 & 48.3 & - & 50.4 & 51.8 & 54.0 & 50.7 & - & - & 51.2(51.2)\\
Meosonet-4 & 57.2 & 59.4 & 59.9 & 47.4 & 53.7 & 57.6 & 48.6 & 61.6 & 66.5 & 56.8(54.8)\\
MeosoIncept-4 & 60.0 & 55.8 & 59.6 & 48.3 & 54.5 & 63.0 & 49.8 & 60.5 & 54.7 & 56.2(55.8)\\
Patch-resnet & 57.1 & 55.9 & 54.1 & 54.5 & 60.0 & 64.9 & 57.7 & 59.8 & 56.7 & 57.8(57.7)\\
Patch-xception & 53.3 & 53.5 & 56.4 & 53.2 & 60.3 & 63.4 & 64.0 & 52.8 & 58.3 & 57.2(57.7)\\
Xception & 55.9 & 61.9 & 52.9 & 59.8 & 61.6 & 52.1 & 48.0 & 56.3 & 57.0 & 56.1(56.0)\\
FFD &  53.9 & 63.4 & 62.1 & 64.3 & 57.8 & 57.4 & 43.5 & - & - & 57.4(57.4)\\
Multi-attention& 52.0 & 54.0 & 64.1 & 59.3 & 54.8 & 54.5 & 57.4 & 74.7 & 74.5 & 60.5 (56.5)\\
Conv LSTM & 56.3 & 57.7 & 59.9 & 50.0 & 62.9 & 55.5 & 39.9 & 52.2 & 50.1 & 53.8(54.6)\\
Average AUC\% & 55.3 & 55.8 & 58.5 & 55.3 & 58.2 & 58.5 & 51.7 & 60.4 & 61.3 & -\\
\hline
\end{tabular}
\caption{Results of generalization ability of detection methods measured by frame-level AUC. Models trained by different datasets are evaluated on ID test set. 
The last row calculates the average AUC of detection methods trained on each dataset.}
\label{tab:generalization_ability}
\end{center}
\end{table*}

\subsection{Evaluation Results of Generalization Ability}
\label{Evaluation Results of Generalization Ability}
In this section, different popular algorithms are trained on different training datasets and evaluated on our ID test set. Since the data in ID test set is diverse and high-quality enough to simulate the real-world situation, this evaluation enables to reflect the generalization ability of forensic algorithms. It can be seen from Table \ref{tab:generalization_ability}, the forgery detection ability of all 11 popular deepfake detection approaches drops significantly, indicating the existing methods remain far from the expectations for real-world deployment. We attribute this performance drop to the multi-domain distribution and imperceptible artifacts of our challenging ID test set.

It can be seen from the results that Multiple-attention, Patch-Resnet-Layer1, and Patch-Xcpetion-Block2 gain relatively top average AUC scores of 60.5\%(56.5\%), 57.8\%(57.7\%), and 57.2\%(57.7\%). This indicates that different manipulations leave common forgery clues in low-level features and focusing on exploring these features benefit detection generalization. Moreover, Face X-ray and FFD also show relatively outstanding generalization ability, indicating that paying more attention to the pixel level manipulation region also helps to learn more fine-grained and precise features and benefits to the generalization.
From the results of average AUC among different detection methods for datasets, we can conclude that DFDC including 6 types of manipulation helps generalize to IDtest set. 

\begin{figure}
\begin{center}
   \includegraphics[width=\linewidth]{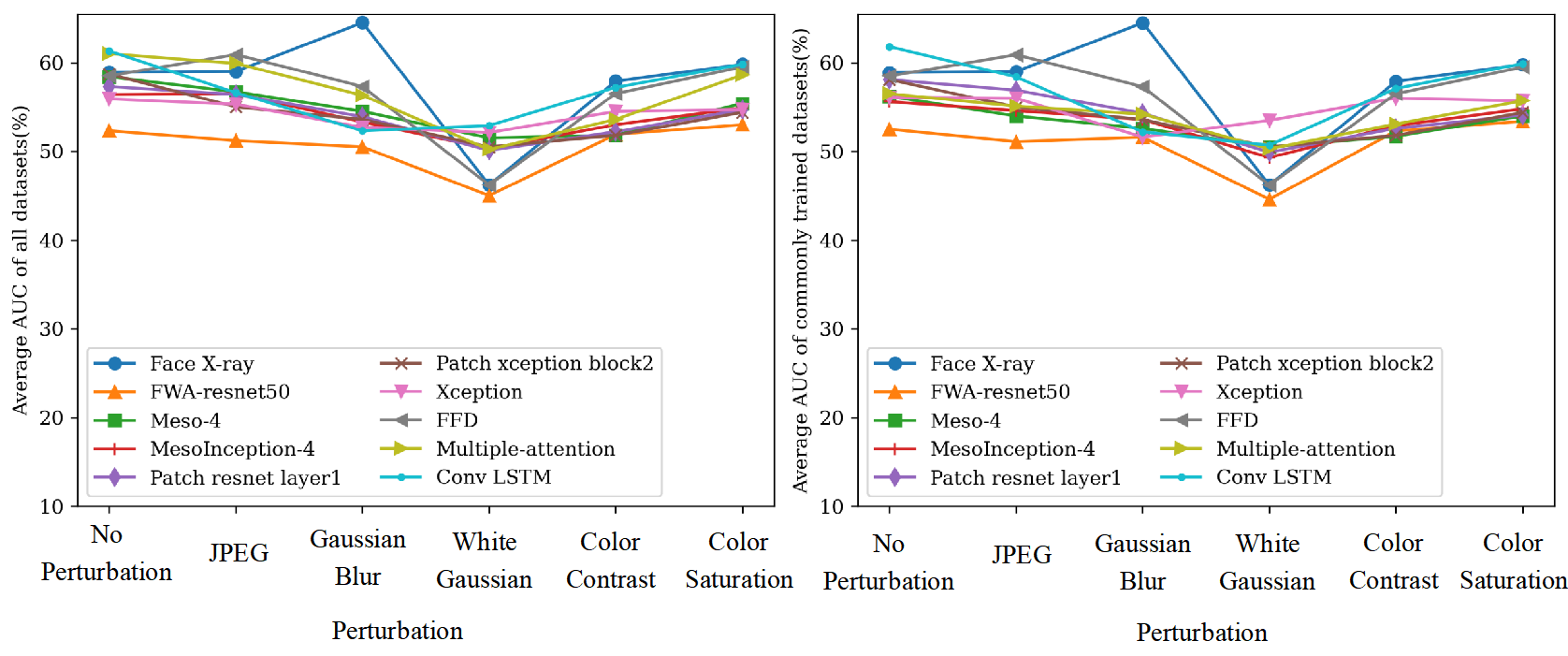}
\end{center}
\vspace*{-3mm}
   \caption{Illustration of the impact of different types of perturbation on average AUC.}
\label{fig:perturbation_auc}
\label{fig:onecol}
\end{figure}

\begin{figure}[t]
\begin{center}
   \includegraphics[width=\linewidth]{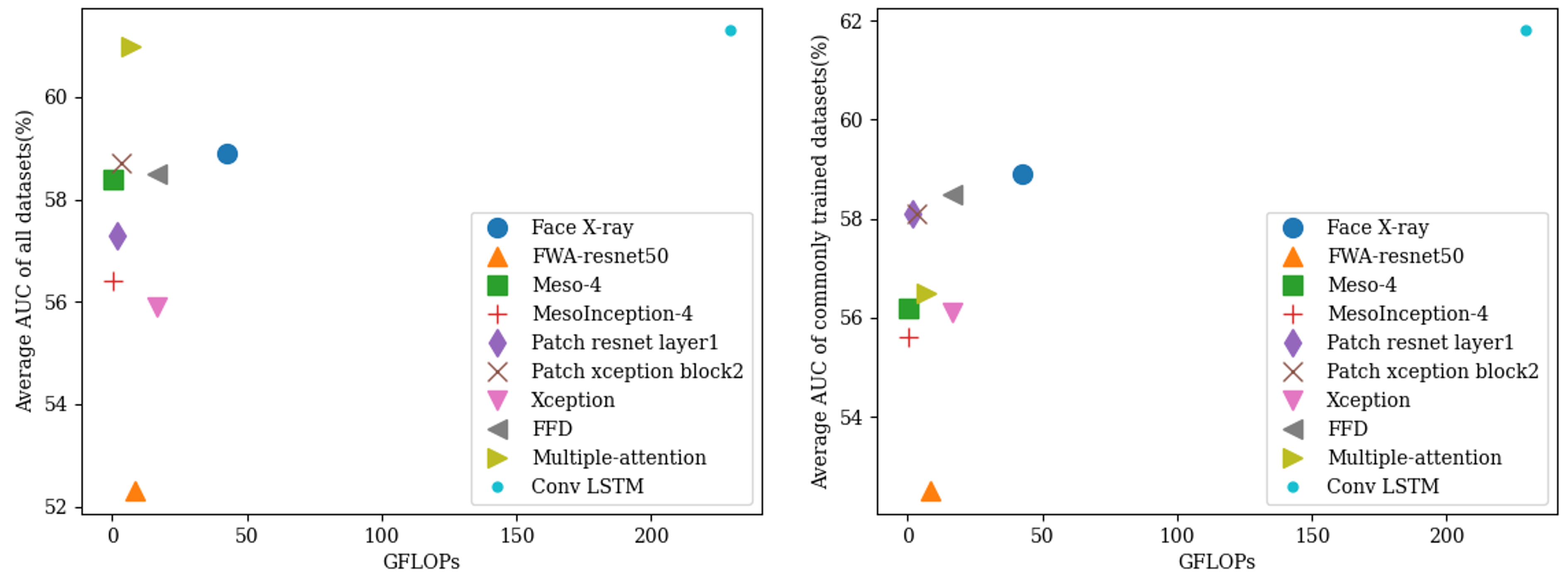}
\end{center}
   \caption{Illustration of the relationship of FLOPs and average AUC score evaluated on ID test set.}
\label{fig:FLOPs_auc}
\label{fig:onecol}
\end{figure}

\subsection{Evaluation Results of Detection Robustness}

Detection robustness is reported in this section on our perturbed ID test set. As shown in Figure~\ref{fig:perturbation_auc}, we report the average detection performance of each algorithm when facing five kinds of perturbations. The average AUC in these two figures are that across all datasets and that across commonly trained datasets. From the results, the detection performance of all methods has experienced varying levels of fluctuations, while the white Gaussian noise can cause a significant degradation of detection performance of most models. 
Among plotting perturbation versus AUC curves, the curve of Xception is the most smooth one. By calculating the population standard deviation of perturbed average AUCs of each algorithm, Xception gains the smallest value of 1.36(1.69). It indicates that this approach, with acceptable performance, is more robust to different kinds of perturbations. Moreover, Mesonet-4 and MesoInception-4 also gain relatively small value of 2.47(1.83) and 2.15(2.05). 


\subsection{Evaluation Results of Practicability}

\begin{figure}[t]
\begin{center}
  \includegraphics[width=\linewidth]{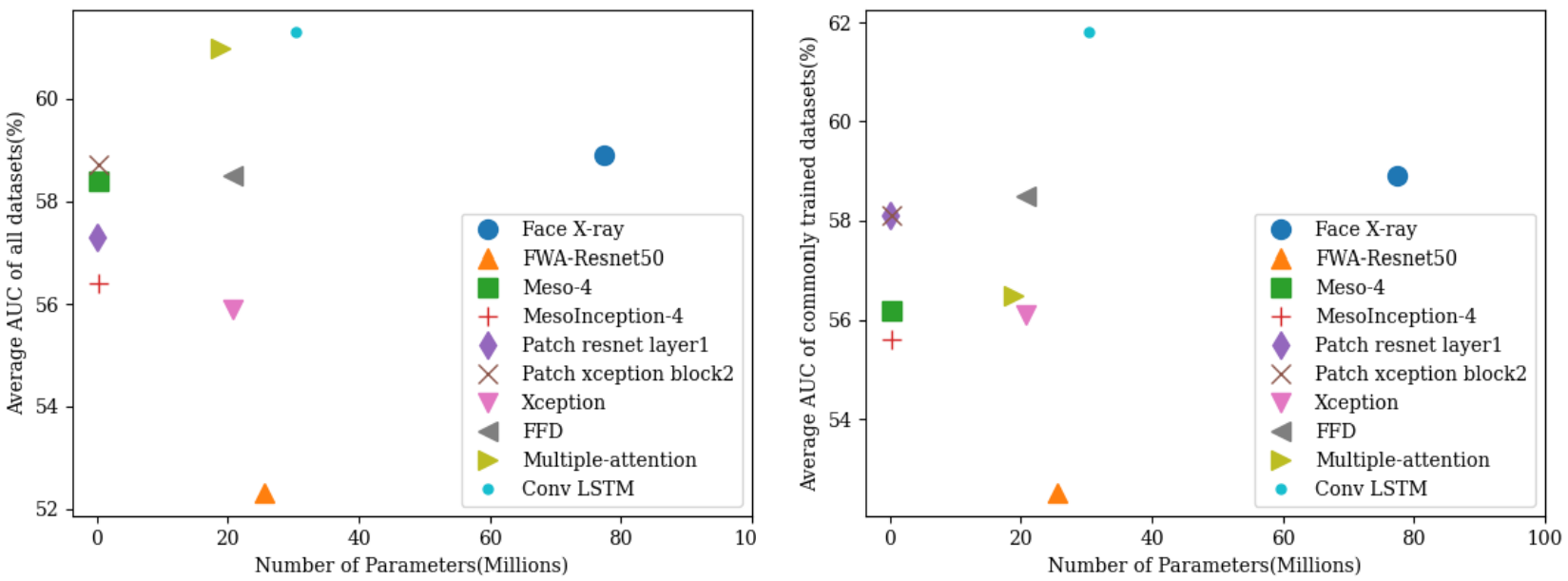}
\end{center}
  \caption{Illustration of the relationship of number of parameters and average AUC score evaluated on ID test set.}
  \vspace*{0mm}
\label{fig:param_auc}
\label{fig:onecol}
\end{figure}

We assess the practicability of forgery detection methods by analyzing AUC versus FLOPs curve and AUC versus number of parameters curve in this section, as shown in Figure~\ref{fig:FLOPs_auc} and Figure~\ref{fig:param_auc}. 
As illustrated in these plots, we can observe that the lightweight Patch-xception-block2 model possesses an acceptable and stable performance, which can be more suitable for practical scenes. 



\subsection{Evaluation Results of Efficiency/Effectiveness Trade-off}

\begin{figure}
\begin{center}
  \includegraphics[width=\linewidth]{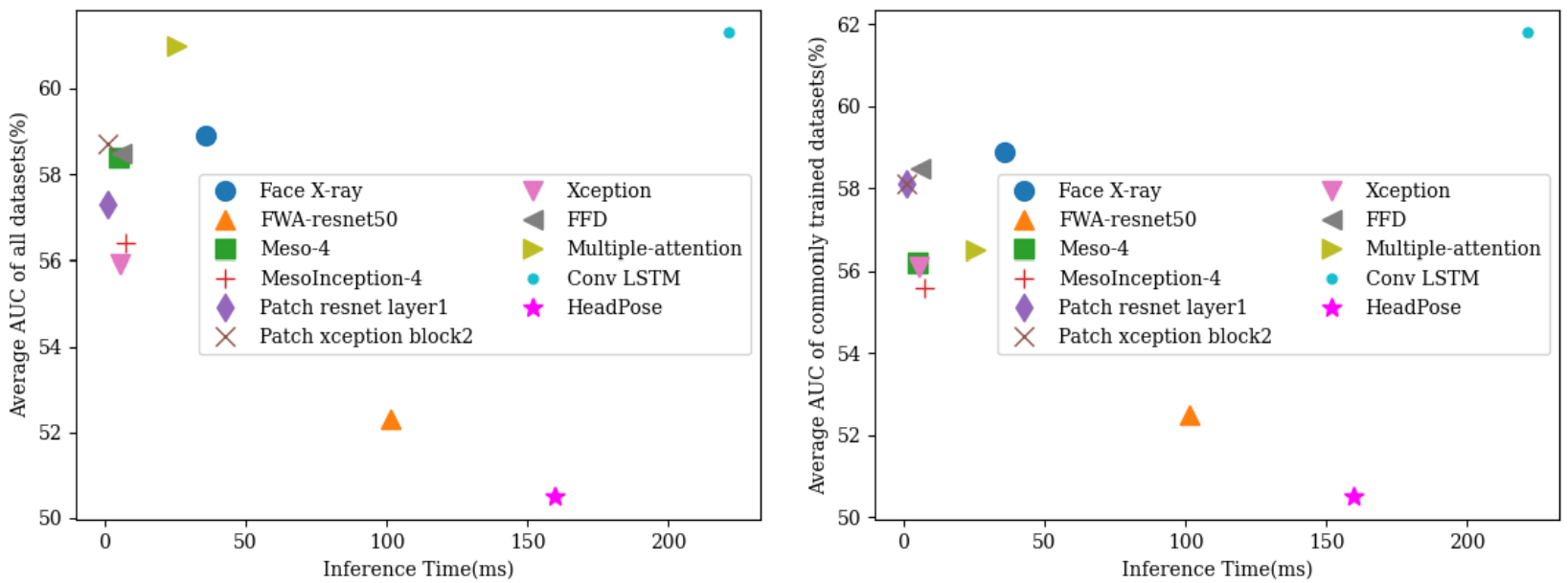}
\end{center}
  \caption{Illustration of the relationship of inference time and average AUC score evaluated on ID test set.}
  \vspace*{0mm}
\label{fig:inference_time_auc}
\label{fig:onecol}
\end{figure}

The evaluation results of efficiency and effectiveness trade-off are shown in Figure~\ref{fig:inference_time_auc}. We measure this metric by exploring the relationship between single-frame inference time of a model and its average AUC score on ID test set. By observing the ratio of AUC to inference time and the quantitative results in Table \ref{tab:detection_methods} and Table \ref{tab:generalization_ability}, we can conclude that Patch-resnet-layer1 achieves relatively superior performance with very low inference time and high AUC score. In addition, as illustrated in Table \ref{tab:detection_methods}, the inference time per frame for most methods is longer than 5 milliseconds, indicating the existing deepfake detection algorithms can be time-consuming and less practical on very large-scale forgery data detection in realistic scenarios.

\section{Conclusion}
In this paper, a comprehensive and consistent benchmark has been established for holistic and fair evaluation of existing deepfake detection approaches. 
By performing large-scale experiments with several fair-minded and practical evaluation metrics, we have concluded that the dataset inconsistencies can lead to unfair comparison among popular approaches.
A challenging ID test set including manipulated samples that are indistinguishable to both humans and detection algorithms, is collected for a better evaluation and understanding of state-of-the-art deepfake detection methods. The evaluation results reveal that the existing popular deepfake detection algorithms remain far from the expectations for real-world deployment. 
The evaluation from multiple perspectives indicates different algorithms have their own advantages and no one method shows comprehensively superiority over others.



{\small
\bibliographystyle{ieee_fullname}
\bibliography{egbib}
}

\section{Appendix A: Evaluation Details}

We summarize our evaluated 11 forgery detection methods (9 algorithms) and the corresponding parameter settings in our experiment in this section. In general, we set the number of training iterations and epochs to an adaptive value according to the quantity relationship between the data scale of our training datasets and that in original papers. And the other specific training parameters are all identical to the original papers. 
For each method, we tune the parameter to achieve the best possible performance.
\subsection{Algorithm Introduction and Implementation Details}
\paragraph{Face X-ray}~\cite{li2020face} is an intra-frame level knowledge-driven detection method. This method applies Face X-ray, the blending boundary produced by the blending procedure in face manipulation, as its interpretable artifacts to guide the model to learn this additional supervised information and then do the classification. The original paper uses \textit{BI} dataset, containing self-supervised learning generated data, as well as public dataset to perform model training. We merely adopt public dataset to train the model in our work for fair evaluation of benchmarking datasets. We adopt HRNet-W48-C~\cite{SunXLW19} as backbone network in the experiment and set batch size to 32. The total number of iterations, numbers of warming start iterations and fine-tuning iterations are set to adaptive values. In addition, the learning rate setting is identical to original paper as 0.0002 using Adam optimizer and it is linearly decayed to 0 for the last adaptive iterations.

\paragraph{FWA}~\cite{li2018exposing} is classified as an intra-frame level knowledge-driven method. Exploring the artifacts caused by affine transform procedure in deepfake production pipeline, this method leverages the resolution inconsistency between manipulated face region and its surrounding region as interpretable artifact. 
To impose detection model to focus on this artifact, this method adopts self-supervised learning to generate negative examples for training and we implement this strategy on the half training positive examples of each dataset in our data preprocessing step. We use ResNet50 as detection model and set batch size as 64. For the number of learning rate decay steps, fine-tuning epochs and hard mining epochs, we set them to adaptive values. Additionally, following the original experimental setup, learning rate respectively starts from 0.001 in fine-tuning stage and starts from 0.0001 in hard mining stage, and decay rate is 0.95.

\paragraph{HeadPose}~\cite{yang2019exposing} is considered as an intra-frame level knowledge-driven method. It uses inconsistent head poses estimated between facial landmarks of the whole face and that of the central face region as interpretable artifact. And then it adopts the SVM classifier to distinguish this difference for deepfake detection. 

\paragraph{Mesonet-4/MesoInception-4}~\cite{afchar2018mesonet} is an intra-frame level data-driven method. By analyzing images at a mesoscopic level, this method introduces two networks, Meosonet-4 and MesoInception-4. Mesonet-4 is a shallow network with a sequence of four layers of successive convolutions and pooling, and followed by a dense network with one hidden layer. MesoInception-4 has a similar network structure which replaces the first two convolutional layers by the inception module with dilated convolutions. 
In the experiment, we set batch size to 75 and use Adam optimizer in the training process. The learning rate starts from \(10^{-3}\) and is divided by 10 every adaptive iterations down to \(10^{-6}\).

\paragraph{Patch Resnet Layer1/Patch Xception Block2}~\cite{chai2020makes} belongs to the intra-frame level data-driven method. By truncating Resnet and Xception after intermediate blocks, this method analyzes images based on patch level predictions. In our experiment, we adopt patch level labels and predictions to calculate losses and metrics to force the model to learn local features in training phase. While in test phase, we aggregate the patch level predictions in an average manner to get the image level predictions and then calculate the image level losses and metrics. Following the original paper, we set batch size as 32, consisting 16 real and 16 fake images, and use Adam optimizer with default parameters and learning rate.

\paragraph{Xception}~\cite{rossler2019faceforensics++} is classified as an intra-frame level data-driven method. This method adopts XceptionNet~\cite{chollet2017xception} as backbone to implement a binary classification. Similar as the original paper, we set batch size as 32 and train the network with a learning rate of 0.0002 using Adam optimizer. We stop the training process when the validation AUC does not change for 10 consecutive checks.

\paragraph{FFD}~\cite{dang2020detection} can be considered as an intra-frame level knowledge-driven method. This method applies an attention mechanism to detect and localize manipulation regions. The author proposes two types of attention-based layer, named manipulation appearance model and direct regression, to guide the network to focus on discriminative regions. 
Meanwhile, three types of loss function are proposed to supervise the learning progress. In our implementation, we adopt the XceptionNet~\cite{chollet2017xception} as the backbone and direct regression as the attention-based layer to train the model.

\paragraph{Multiple-attention}~\cite{zhao2021multi} is an intra-frame level knowledge-driven method, which considers deepfake detection as a fine-grained classification problem and proposes a multi-attentional deepfake detection network. 
This network applies an attention module to generate multiple attention maps assisting explore local discrimination, utilizes densely connected convolutional layers to enhance subtle texture artifacts in shallow feature map, and uses bilinear attention pooling to aggregate the low-level textural feature and high-level semantic features guided by the attention maps.  
What's more, this method designs a regional independent loss to learn multiple attention maps and applies AGDA mechanism to force the attention to mine more useful information. Following the original paper, we use EfficientNet-b4~\cite{tan2019efficientnet} as the backbone network and set the \textit{SL$_{t}$} and \textit{SL$_{a}$} as L2 and L5.

\paragraph{Conv LSTM}~\cite{guera2018deepfake} is an inter-frame level detection method. It takes consecutive frame images as input and designs a network with InceptionV3 to extract frame level features and uses LSTM to detect the anomaly between frames. To achieve the best performance, instead of directly feeding consecutive frame images into network in original paper, we implement face detection in data pre-processing and use consecutive face images as network input. We train our model using a sequence of 20 images and a batch size of 4. The optimizer is set to Adam with an initialized learning rate of 1e-5.
\begin{table*}
\begin{center}
\begin{tabular}{l@{\hspace{5pt}}c@{\hspace{5pt}}c@{\hspace{5pt}}c@{\hspace{5pt}}c@{\hspace{5pt}}c@{\hspace{5pt}}c}
\hline
\multirow{2}{*}{Method}  &  
\multicolumn{3}{c}{Reported Result} & 
\multicolumn{3}{c}{Our Result}\\
& Test Dataset & Metric & Result & Test Dataset & Metric & Result \\
\hline\hline
\multirow{2}{*}{Face X-ray} & FF++/DF & AUC & 99.17~\cite{li2020face} & FF++/DF & AUC & 99.4\\
& FF++/FS & AUC & 99.20~\cite{li2020face} & FF++/FS & AUC & 99.8\\
\multirow{2}{*}{FWA-resnet} & DeepfakeTIMIT & AUC & 87.4 & DeepfakeTIMIT & AUC & 99.1\\
&UADFV & AUC & 79.0 & UADFV & AUC & 57.3\\
MeosoInception-4 & paper released data & ACC& 91.7~\cite{afchar2018mesonet} & paper released data & ACC & 91.2  \\
Xception & FF++/DF & ACC & 99.59~\cite{rossler2019faceforensics++} & FF++/DF & ACC & 96.7 \\
Multiple-attention & FF++ & AUC & 99.29~\cite{zhao2021multi} & FF++/DF, FF++/FS & AUC & 99.4 \\
Conv LSTM & Unavailable data & ACC & 96.7~\cite{guera2018deepfake} & FF++/DF & ACC & 97.9\\
\hline
\end{tabular}
\vspace{2pt}
\caption{Results of evaluation of re-implementation correctness.}
\label{tab:implementation_correctness_evaluation}
\end{center}
\end{table*}
\subsection{Data Pre-processing Pipeline} 
We apply the image processing toolkit Dlib~\cite{king2009dlib} and OpenCV-Python~\cite{opencv_library} to implement the common data pre-processing, including frame extraction, face cropping, face detection and face alignment. Here we specially describe the additional data pre-processing pipeline in details for the algorithms requiring those processes but without corresponding open-sourced codes.

\paragraph{Face X-ray} requires the additional mask data, named Face X-ray to supervise the learning process. It defines Face X-ray as an image I with 
\begin{equation}
    I_{i,j} = 4 \cdot M_{i,j} \cdot (1 - M_{i,j})
\end{equation}
where M is the soft non-binary mask which delimits the manipulated region. Following the paper, we generate the ground-truth Face X-ray for fake images. Firstly, the binary difference mask can be generated by computing the absolute element-wise difference between manipulated image and its mapping swapped target image. Then a Gaussian blur followed by normalization is applied to generate soft non-binary mask. Finally, we use the above formulation to generate Face X-ray.

\paragraph{FWA} adopts self-supervised learning to generate negative examples in the model training phase. Following its paper, we firstly align the face image to multiple scales and randomly select an aligned image. Then we apply Gaussian blur to the aligned image and affine warp it back to the original size. Finally, we randomly change the image color information and paste the whole face region or landmark region to the original frame image to get the negative example.

\paragraph{FFD} requires the ground-truth binary modification mask to get the attention map loss. Following the paper, we compute the absolute pixel-wise difference of RGB image pairs between manipulated images and their corresponding source images. Then we convert them to grayscale images and attain the binary masks by thresholding the normalized grayscale images. 

\section{Appendix B: Evaluation of Re-Implementation Correctness}
To ensure the correctness of our re-implemented detection algorithms, we have conducted verification experiments of those methods not or partially open-sourced to show the consistency between our results and the reported results. The comparative results are shown in Table~\ref{tab:implementation_correctness_evaluation}. 

For Conv LSTM, since its test data used in original paper is unavailable, we use FF++/DF to test the model and show that we can get a better result.
For all the other methods, we utilize the same domain distributed data to execute the verification.
Since the unavailability of totally same frame data, we randomly extract frame data from each test dataset for fair comparison. 
It can be seen from the table, for Face X-ray, FWA-resnet (DeepfakeTIMIT), MesoInception-4, Multiple-attention and Conv LSTM approaches, our re-implementation can achieve better or comparable results. 
While the results of FWA-resnet and Xception exists some gaps of the reported results.
For FWA-resnet, we speculate that such approach strongly relies on the image quality of the generated training samples. We find it is easy to generate high quality negative examples based on DeepfakeTIMIT. Therefore, we can achieve superior performance on this dataset. And the low quality negative examples generated based on UADFV lead to poor performance of this dataset.
For Xception, we are unable to get the superior result reported in the original paper. A similar claim that they only manage to fine-tune Xception to obtain a accuracy score of 96.1\% has also been mentioned in ~\cite{afchar2018mesonet}. 

\end{document}